\pdfoutput=1
\documentclass[11pt]{article}

\usepackage[preprint]{acl}

\usepackage{times}
\usepackage{latexsym}
\usepackage[T1]{fontenc}
\usepackage[utf8]{inputenc}
\usepackage{microtype}
\usepackage{inconsolata}
\usepackage{graphicx}
\usepackage{booktabs}
\usepackage{amsmath}
\usepackage{array}
\newcolumntype{L}[1]{>{\raggedright\arraybackslash}p{#1}}

\title{Calibrated to Whom? Persona and Language Effects\\on Cultural Values in JEV}

\author{
  \textbf{Bushra Asseri}\textsuperscript{1,2} \qquad \textbf{Abdulaziz Asseri}\textsuperscript{2} \\
  \textsuperscript{1}College of Engineering and Advanced Computing, Alfaisal University, Riyadh, Saudi Arabia \\
  \textsuperscript{2}Proxa.sa, Riyadh, Saudi Arabia \\
  \href{mailto:basseri@alfaisal.edu}{\texttt{basseri@alfaisal.edu}}, \href{mailto:abdulaziz.asseri@gmail.com}{\texttt{abdulaziz.asseri@gmail.com}}}

\begin{document}
\emergencystretch=1.5em
\raggedbottom
\maketitle
\begin{abstract}
Decision-only language models return a probability for every answer option instead of generating text, which makes them attractive as survey respondents and as judges. We audit the cultural values of one such model, TypeSafe's JEV, with the Values Survey Module 2013. We asked it the 24 items as 12 matched Saudi and 12 matched American personas and without a persona, in English and Arabic, under eight ways of formulating the request (288,000 answers). JEV's answers were highly repeatable (ICC 0.997). Without a persona, they were closer to those of its own American personas than to those of its Saudi personas, though distinct from both. When the persona was Saudi rather than American, the answers moved in the direction of the Saudi--US difference in Hofstede's scores, which we use as a reference benchmark. Along that direction they covered 87\% of its size in English and 62\% in Arabic, but their overall discrepancy from it was about 60\% of its size in both languages, mostly because long-term orientation was reversed. A language cross associates the smaller Arabic difference mainly with the language of the items rather than of the persona description. Age shifted the profiles about as much as nationality, gender shifted them more for Saudi than for American personas, and in the documented request design JEV was less confident in Arabic. The national difference kept its direction in all eight designs: patterns known from generative models appear in a model that never generates text.
\end{abstract}

\section{Introduction}

\begin{figure}[t]
\centering
\includegraphics[width=\columnwidth]{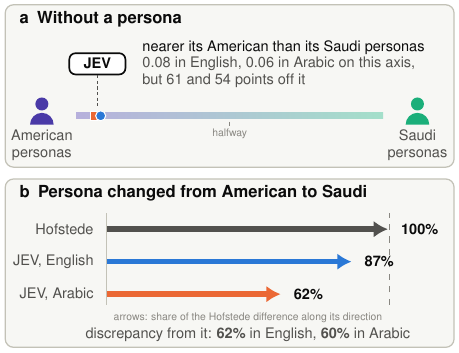}
\caption{Calibrated to whom? (a) Without a persona, JEV's answers lie nearer its own American than its Saudi personas on the axis between them (0 = American, 1 = Saudi personas), but well off that axis. (b) When the persona changes from American to Saudi, the answers move 87\% (English) and 62\% (Arabic) of the Saudi--US difference in Hofstede's scores along its direction, while their full difference departs from Hofstede's by 62\% and 60\% of its size (documented set-up).}
\label{fig:teaser}
\end{figure}

\begin{figure*}[t]
\centering
\includegraphics[width=\textwidth]{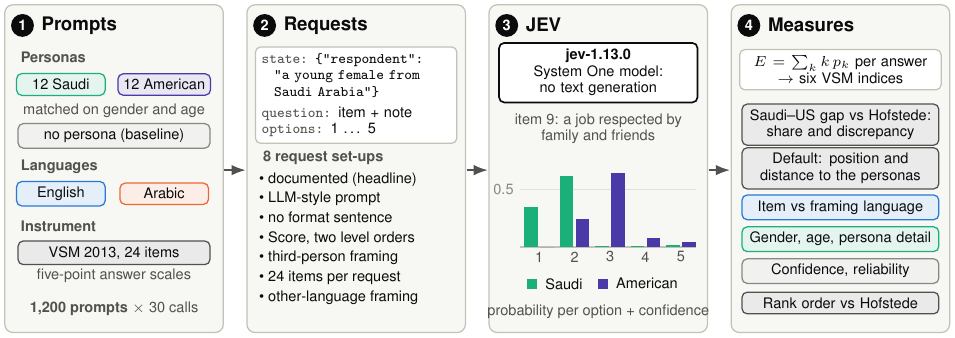}
\caption{Study design. (1) Matched Saudi and American personas and a no-persona baseline, in English and Arabic, answer the 24 VSM 2013 items. (2) Each prompt becomes a JEV request in eight set-ups. (3) JEV returns a probability for each answer option; the bars show its mean probabilities for item 9 (``how important would it be to you to have a job respected by your family and friends'') for a young Saudi and a young American woman (documented set-up, English; 1 = of utmost importance). (4) Expected answers are scored on the six VSM dimensions, and JEV's profiles are compared with each other and with the Saudi--US difference in Hofstede's scores.}
\label{fig:overview}
\end{figure*}

Value surveys have become a standard way to ask whose values a language model reflects \citep{santurkar2023whose,durmus2023towards,cao2023assessing,alkhamissi2024investigating,tao2024cultural}. With generative models the procedure is indirect: many completions are sampled, free text is mapped back to answer options, refusals have to be handled, and the results move with the format of the prompt \citep{rottger2024political}.

A new class of models removes these steps. JEV, released by TypeSafe AI in September 2026, is described as a ``System One'' model: given a context and a question with fixed options, it returns a probability for every option and never generates text \citep{typesafe2026}. Within weeks it was evaluated as an inexpensive first-pass judge \citep{li2026jevjudge} and used to predict the answers of described respondents in population experiments \citep{li2026kite}. Both studies worked in English, and neither asked whose values the model itself reports.

For a model that stands in for respondents, three questions matter. What does it answer when no respondent is described? When the respondent's nationality is given, do its answers move in the direction of published national differences, and how far? And do the answers depend on the language of the questionnaire? Figure~\ref{fig:teaser} previews our answers. The language can matter even for a single item: asked in English whether an organisation's rules should never be broken, JEV's Saudi personas agree more strongly than its American personas, whereas in Arabic the two answer alike. Arabic makes the last question pressing: cultural biases in Arabic contexts are well documented for generative models \citep{naous2024having,alkhamissi2024investigating}, and JEV is trained mainly on English, with lower documented accuracy in other languages \citep{typesafe2026}.

We answer these questions with the Values Survey Module 2013 \citep[VSM 2013;][]{hofstede2013vsm}. We describe 12 matched Saudi and 12 matched American personas, or none, in English and in Arabic, and compare the Saudi--American difference in JEV's answers with the difference between the two countries' published Hofstede scores, which serves as a reference benchmark rather than as a matched human sample (Figure~\ref{fig:overview}). Because the absolute position of VSM scores depends on constants that are not published for the reference countries \citep{bulte2026llms}, every measure compares two of JEV's own profiles, so that the constants cancel. Each of the 1,200 prompts was sent 30 times under eight ways of formulating the request, 288,000 answers in total.

Our contributions are:
\begin{itemize}
\setlength{\itemsep}{1pt}
\item the first audit of the cultural values of a decision-only model, in English and Arabic, with constant-free measures of the national difference and of the default profile;
\item evidence that JEV's answers follow the direction of the reference Saudi--US difference, covering 87\% of its size along that direction in English and 62\% in Arabic while departing from it by about 60\% of its size in both languages; that the smaller Arabic difference is associated mainly with the language of the items rather than of the persona description; and that age and gender shift the profiles in stereotyped ways;
\item a robustness analysis over eight request designs that separates findings that depend on how the model is asked from those that do not.
\end{itemize}

\section{Related Work}

\paragraph{Survey-based cultural evaluation.}
Studies of cultural values in language models administer value surveys and compare the answers with national results, using Hofstede's dimensions \citep{cao2023assessing,masoud2025cultural} or the World Values Survey and related opinion polls \citep{santurkar2023whose,durmus2023towards,alkhamissi2024investigating,tao2024cultural}. Recurring findings are that answers resemble those of the United States or other Western countries \citep{cao2023assessing,durmus2023towards,tao2024cultural}, also when the question is asked in another language \citep{wang2024not}; that naming a country or using its language moves the answers only partway towards it \citep{durmus2023towards,alkhamissi2024investigating,tao2024cultural}; and that answers given for a group can reproduce stereotypes \citep{durmus2023towards}. In Arabic contexts, models favour Western entities and norms \citep{naous2024having}, and GPT-4's answers to the VSM 2013 matched Chinese values better than American or Arab ones \citep{masoud2025cultural}. We use the same family of instruments, but for a model that does not generate text.

\paragraph{Validity of survey measurement.}
Forced-choice answers from language models can differ from their unconstrained responses and change with small variations of the prompt \citep{rottger2024political}. For the VSM, each dimension index contains a constant chosen by the researcher, so the distance between a model's scores and published country scores is not identified \citep{bulte2026llms}. The VSM manual adds that scores are comparable only between samples matched on everything but nationality, and that new scores are not comparable to the published ones \citep{hofstede2013vsm}. Lacking the constants, \citet{masoud2025cultural} compared the order of countries on each dimension with the order of their published scores instead. We therefore compare profiles measured on the same scale and vary the request design systematically.

\paragraph{Language models as respondents.}
Conditioning a model on demographic descriptions has been proposed as a way to simulate survey samples \citep{argyle2023out}, while other work shows that model opinions match some groups better than others and that steering towards a group succeeds only in part \citep{santurkar2023whose}.

\paragraph{JEV.}
JEV is a proprietary model released by TypeSafe AI in September 2026 as the first of what the company calls ``System One'' models, named after the fast, intuitive mode of thinking described by \citet{kahneman2011thinking} \citep{typesafe2026}. It does not generate text. Given a context, the \emph{state}, it answers typed questions: a Choice question returns a probability for each of up to 255 named options, a Score question a probability for each of two to ten ordered levels, and a yes/no question the probability of yes. All probabilities are produced in a single non-autoregressive pass, and each answer carries a confidence value. TypeSafe trains the model with what it calls reinforcement learning for calibrated decisions, positions it as a fast and inexpensive decision component for software, and notes that it is trained mainly on English and is less accurate in other languages. \citet{li2026jevjudge} found it within three points of a frontier judge on preference and factuality judgements, and weaker where a judgement requires checking a derivation, at a fraction of a percent of the cost. \citet{li2026kite} used it to predict how described respondents would answer in population experiments, with English text only. Neither study examined cultural values or Arabic.

\section{Experimental Setup}\label{sec:setup}

\subsection{The Model}

\paragraph{Typed questions and answers.}
We used two of JEV's question types. A \emph{Choice} question lists named options with descriptions; its answer gives the most probable option, a probability for every option and a confidence value between 0 and 1 derived from the shape of the distribution. A \emph{Score} question lists ordered levels and returns a probability for each. The state can be text or a JSON object, questions in one request are answered independently, and the model has no temperature or seed parameter \citep{typesafe2026}.

\paragraph{Access.}
We sent all requests to TypeSafe's API and asked explicitly for version \texttt{jev-1.13.0}, which every response reported. The data were collected on 28 and 29 September 2026. Probabilities are returned with two decimals; we renormalised them to sum to one. Cost and software are listed in Appendix~\ref{app:repro}.

\subsection{Instrument and Reference Values}

\paragraph{VSM 2013.}
We used the 24 content items of the VSM 2013, each answered on a five-point scale, and computed the six dimension indices (power distance, individualism, masculinity, uncertainty avoidance, long-term orientation, indulgence) with the manual's formulas (Appendix~\ref{app:vsm}).

\paragraph{Reference values.}
We used Hofstede's country scores for Saudi Arabia, as published by \citet{hofstedeinsights2022}, and for the United States \citep{hofstede2010cultures}; as an alternative Saudi profile, we used the VSM 2013 scores of \citet{almutairi2021reclaiming}. These scores come from different respondents, periods and survey versions, and the VSM manual warns that scores are comparable only between matched samples \citep{hofstede2013vsm}. We therefore use the Saudi-minus-US difference in Hofstede's scores, $\mathbf h$ (length 94.3; Appendix~\ref{app:vsm}), as a reference benchmark for the direction and approximate size of the national difference, not as a target that synthetic personas should match exactly.

\paragraph{Constant-free comparisons.}
Each index contains an additive constant chosen by the researcher, and the constants behind published country scores are not reported, so the absolute distance between a model's indices and a country's scores is not identified \citep{bulte2026llms}. All our measures compare two profiles on the same scale, so the constants cancel. Differencing removes the constants, but not the differences in samples, periods and versions noted above.

\subsection{Personas, Prompts and Languages}

\paragraph{Matched personas.}
For each country, a persona names the country alone, the country with a gender (female, male) or an age group (young, middle-aged, elder), or the country with both. The resulting 12 demographic profiles are the same for Saudi Arabia and the United States, so Saudi and American personas can be compared in matched pairs. A baseline asks the items without a persona.

\paragraph{Prompts.}
English items follow the official VSM 2013 wording, recast as questions (``In choosing an ideal job, how important would it be to you to have sufficient time for your personal or home life?''); Arabic items follow the official Arabic translation \citep{vsm2013arabic}, recast in the same way.

\paragraph{Repetitions.}
With 25 persona conditions, two languages and 24 items there are 1,200 prompts. Each was sent 30 times in each of the eight set-ups, giving 36,000 answers per set-up.

\subsection{Request Set-ups}\label{sec:setups}

JEV has no system and user roles, so a persona prompt can be passed to it in several ways (Table~\ref{tab:setups}; texts in Appendix~\ref{app:prompts}).

\paragraph{Documented set-up.}
Our headline set-up follows TypeSafe's recommendation to give content as named fields of the state and the task in the question. The state is \texttt{\{"respondent": "a young female from Saudi Arabia"\}}, and a Choice question contains the item wording with the note ``Answer as the respondent described in `respondent` would answer'' and the five answer labels as options.

\paragraph{LLM-style prompt and variations.}
Set-up 2 uses a persona prompt in the form common for chat models: the system prompt (``Imagine you are a young female from Saudi Arabia. Answer the following question from your perspective as a person from this cultural background.'', followed by a format sentence) as the state, and the user prompt, with the numbered options and the item, as the question. Set-ups 3 to 7 each change one element: they drop the format sentence, use a Score question with the levels in questionnaire or reverse order, describe the respondent in the third person and ask which answer they would choose, as in population experiments with JEV \citep{li2026kite}, or send the 24 items of a persona as 24 questions in one request.

\paragraph{Language cross.}
Set-up 8 takes the system prompt from the other language, so that English framing meets Arabic items and Arabic framing meets English items.

\begin{table}[t]
\centering\small
\begin{tabular}{@{}lL{2.0cm}L{4.35cm}@{}}
\toprule
 & Set-up & Request \\
\midrule
1 & Documented (headline) & Persona as a field of the state; item wording and a note; Choice \\
2 & LLM-style prompt & System prompt as state; user prompt as question; Choice \\
3 & No format sentence & As 2, without ``IMPORTANT: Respond with ONLY \ldots'' \\
4 & Score question & As 2, options as ordered levels \\
5 & Score, reversed & As 4, levels in reverse order \\
6 & Third person & ``Description of the survey respondent: [persona].''; ``Which answer would they choose?'' \\
7 & 24 items per call & As 2, 24 questions per request \\
8 & Other-language framing & As 2, system prompt of the other language \\
\bottomrule
\end{tabular}
\caption{The eight request set-ups. Each covers the 1,200 prompts, each sent 30 times.}
\label{tab:setups}
\end{table}

\subsection{Measures}\label{sec:measures}

\paragraph{Answers and profiles.}
For each call we computed the expected answer $E=\sum_k k\,p_k$ from the probabilities $p_1,\dots,p_5$, averaged $E$ over the 30 calls of each prompt and computed the six indices for every persona condition and language.

\paragraph{National difference.}
For each of the 12 demographic profiles $i$ and each language, $\mathbf d_i$ is the index vector of the Saudi persona minus that of the American persona. From the mean $\bar{\mathbf d}$ we computed the share of the reference difference covered along its direction, the direction cosine and the relative discrepancy,
\[
s=\frac{\bar{\mathbf d}\cdot\mathbf h}{\lVert\mathbf h\rVert^{2}},\qquad
\cos=\frac{\bar{\mathbf d}\cdot\mathbf h}{\lVert\bar{\mathbf d}\rVert\,\lVert\mathbf h\rVert},
\]
\[
D=\frac{\lVert\bar{\mathbf d}-\mathbf h\rVert}{\lVert\mathbf h\rVert}.
\]
A share of 100\% is a difference of the reference size along the reference direction. The share ignores deviations in other directions; the discrepancy counts them: $D=0$ reproduces the reference difference exactly, and $D=100\%$ is as far from it as no national difference at all.

\paragraph{Default position.}
The no-persona profile $\mathbf b$ is placed between JEV's own American and Saudi personas, with mean profiles $\bar{\mathbf u}$ and $\bar{\mathbf s}$, by $t=(\mathbf b-\bar{\mathbf u})\cdot(\bar{\mathbf s}-\bar{\mathbf u})/\lVert\bar{\mathbf s}-\bar{\mathbf u}\rVert^{2}$: 0 means that the projection of the default falls on the American personas, 1 on the Saudi personas. Because $t$ only locates $\mathbf b$ along this axis, we also report the distances from $\mathbf b$ to $\bar{\mathbf u}$ and to $\bar{\mathbf s}$ and its distance from the axis, in index points.

\paragraph{Gender, age and detail.}
Female-minus-male differences compare personas matched on country, language and age (four pairs per country and language); elder-minus-young differences are matched on gender (three pairs). The share $s$ is also computed separately for country-only personas and for personas with one or two further attributes.

\paragraph{Confidence and reliability.}
Confidence is averaged over the 30 calls of each prompt and compared across matched prompts. Because matched prompts share personas and items, we test confidence differences on persona means and on item means and report a two-way cluster bootstrap interval that resamples both. Test--retest reliability is the one-way intraclass correlation of $E$ across the 30 calls.

\paragraph{Rank agreement.}
We also applied the rank test of \citet{masoud2025cultural}, which asks whether a model orders countries on each dimension as their published scores do, using Kendall's $\tau$ and the share of dimensions out of order. With two countries, the order on a dimension is the sign of the Saudi-minus-US difference, so $\tau$ is $+1$ or $-1$ for each dimension; we report the number of dimensions in the reference order, $k$, and the mean $\tau=(2k-6)/6$. As in their model-level comparison, we also let the language stand for the country and compared the no-persona profile in Arabic with that in English. The constants do not affect the order of two countries on a dimension.

\subsection{Statistical Analysis}

Repeated calls return nearly identical answers and are not independent observations, so matched personas are the units of replication. Intervals for dimension differences are 95\% $t$-intervals across matched profiles or pairs; intervals for shares, discrepancies and their differences are percentile intervals from 4,000 bootstrap resamples of the 12 matched profiles. Confidence differences are tested with Wilcoxon signed-rank tests over persona means and over item means. These intervals describe variation across our constructed persona profiles and items. The profiles are not a sample of people or of possible persona wordings, and the intervals do not include uncertainty in the translation or in the reference scores. The analysis is exploratory and was not pre-registered; as a safeguard, every finding was re-estimated in all eight set-ups.

\section{Results}\label{sec:results}

Unless stated otherwise, results are for the documented set-up.

\begin{figure}[t]
\centering
\includegraphics[width=\columnwidth]{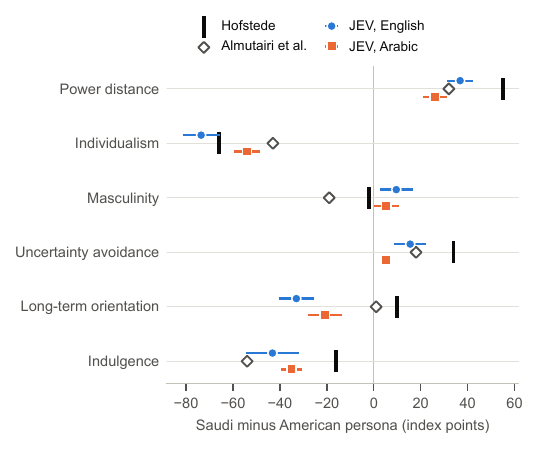}
\caption{Saudi minus American persona on the six VSM dimensions (documented set-up; means over the 12 matched profiles with 95\% CIs), with the reference differences from Hofstede's scores and from \citet{almutairi2021reclaiming}.}
\label{fig:national}
\end{figure}

\begin{figure*}[t]
\centering
\includegraphics[width=\textwidth]{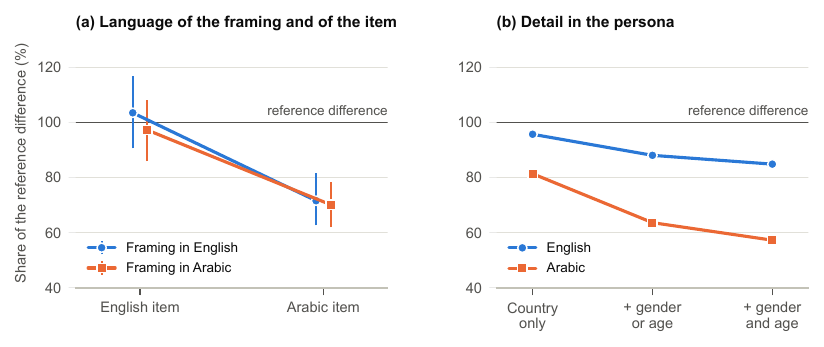}
\caption{Share of the reference (Hofstede) Saudi--US difference covered along its direction. (a) By the language of the framing and of the item (LLM-style prompt and its language cross; 95\% bootstrap intervals over matched profiles). (b) By the detail in the persona (documented set-up): country only (one matched pair per language), one further attribute (five pairs) and two (six pairs).}
\label{fig:language}
\end{figure*}

\subsection{Reliability}
JEV's answers were highly repeatable. The intraclass correlation of the expected answer over 30 calls was 0.997 in the documented set-up and between 0.995 and 0.998 in the others, the standard deviation across calls averaged 0.03 on the five-point scale, and the most probable option was the same in all 30 calls for 83--89\% of prompts (Appendix~\ref{app:reliability}). Mean confidence ranged from 0.49 to 0.58 across set-ups, with the flattest distributions for Score questions.

\subsection{The Default without a Persona}

Without a persona, JEV's answers were closer to those of its own American personas than to those of its Saudi personas, but distinct from both. Projected on the axis from its American (0) to its Saudi (1) personas, the default lay at 0.08 in English and 0.06 in Arabic. In index points, it lay 61.5 from the American and 111.0 from the Saudi centroid in English (54.5 and 87.5 in Arabic), and almost all of its distance from the American centroid lay off that axis (61.0 points, for an axis 100.3 points long; in Arabic 54.3 and 73.0). The default was closer to the American than to the Saudi centroid in 15 of the 16 combinations of set-up and language, with positions between $-0.26$ and 0.13 (Appendix~\ref{app:setups}). The exception was the third-person framing in Arabic, which placed an undescribed respondent midway between the two nationalities (0.53; 50.1 and 48.3 points from the two centroids).

\subsection{The Saudi--American Difference}

\begin{figure*}[t]
\centering
\includegraphics[width=\textwidth]{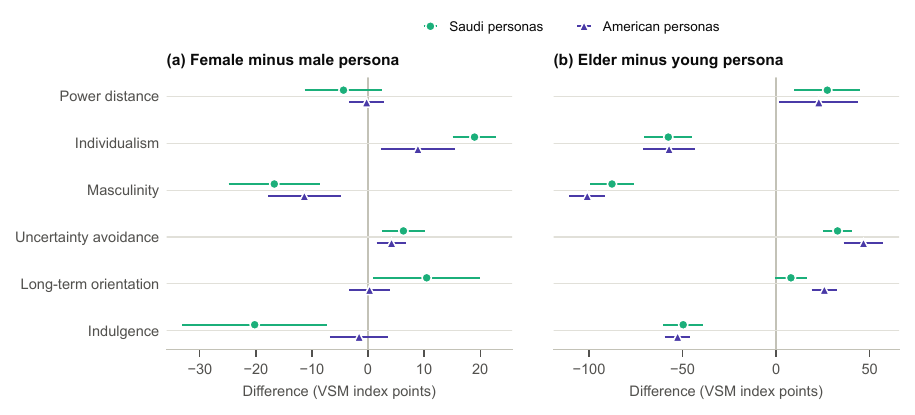}
\caption{Gender (a) and age (b) differences within Saudi and American personas (documented set-up; means over matched pairs in both languages, 8 gender and 6 age pairs per nationality, with 95\% CIs). The two panels use different scales.}
\label{fig:gender}
\end{figure*}

When the persona changed from American to Saudi, JEV's answers moved in the direction of the reference difference (Figure~\ref{fig:national}). In English, its Saudi--American difference covered 87.1\% of the Hofstede difference along that direction (95\% CI 80.5--93.9; cosine 0.82). Its relative discrepancy from the Hofstede difference was nevertheless 62.3\% (58.5--67.6), because most of the deviation lay outside that direction (61.0\% of the reference length, against 12.9\% along it). Against the Saudi profile of \citet{almutairi2021reclaiming}, the share was 104.1\% (94.8--113.6) and the discrepancy 68.7\% (57.4--80.9).

\paragraph{By dimension.}
Four of the five dimensions on which the two countries differ came out in the reference direction: Saudi personas scored higher on power distance ($+36.8$; Hofstede $+55$) and uncertainty avoidance ($+15.5$; $+34$) and lower on individualism ($-73.6$; $-66$) and indulgence ($-43.2$; $-16$), the last nearly three times the reference value. Masculinity, on which the countries barely differ ($-2$), showed a small positive difference ($+9.6$).

\paragraph{Long-term orientation.}
Saudi personas scored 33.0 points lower on long-term orientation ($-40.6$ to $-25.4$), whereas Hofstede's Saudi profile is 10 points higher, the largest single deviation ($-43.0$ points). The VSM index counts greater importance of doing a service to a friend and greater national pride towards short-term orientation, and JEV's Saudi personas gave both: on the 1--5 answer scale, their answers were 0.48 points lower (more important) on the service item in both languages and 0.77 (English) and 0.84 (Arabic) points lower (prouder) on national pride.

\subsection{Language}

In Arabic, JEV covered 62.0\% of the Hofstede difference along its direction (56.3--67.9; cosine 0.80), 25.1 points less than in English ($-31.0$ to $-20.6$). Every dimension kept the sign it had in English, but the gap narrowed on power distance ($+26.0$), individualism ($-54.1$) and uncertainty avoidance ($+5.2$). The deviations shrank with it, so the relative discrepancy was about the same as in English (59.9\%, 55.8--64.2): the Arabic difference was smaller overall rather than less accurate.

\paragraph{Item language rather than framing.}
In this experiment, the language cross associated the drop mainly with the language of the items (Figure~\ref{fig:language}a; LLM-style prompt). With English items, the share was 103.5\% under English framing and 97.2\% under Arabic framing; with Arabic items, it was 71.6\% and 70.2\%. Averaged over framings, an Arabic item lowered the share by 29.5 points (26.1--33.1), whereas an Arabic framing changed it by $-3.9$ points ($-8.9$ to 1.5), which rules out a large framing effect but not a small one.

\paragraph{By item.}
The national difference disappeared in Arabic for two agree--disagree statements, on avoiding two bosses and on never breaking an organisation's rules (English $-0.30$ and $-0.54$ on the answer scale; Arabic $0.00$ and $-0.06$; Appendix~\ref{app:items}).

\paragraph{Persona detail.}
More detail in the persona weakened the national difference in both languages (Figure~\ref{fig:language}b): in English from 95.7\% for country-only personas to 88.1\% with gender or age and 84.9\% with both; in Arabic from 81.4\% to 63.7\% and 57.3\%.

\subsection{Gender and Age}
Female Saudi personas were more individualist ($+19.0$, 15.1--22.9; 8 of 8 matched pairs), less masculine ($-16.7$; 8 of 8) and less indulgent ($-20.2$; 7 of 8) than male Saudi personas (Figure~\ref{fig:gender}). Among American personas the gender pattern was weaker (individualism $+8.9$, masculinity $-11.3$) and absent for indulgence ($-1.6$). Age changed the profiles more, and alike for both nationalities: elder Saudi personas scored higher than young ones on power distance ($+27.4$) and uncertainty avoidance ($+32.9$) and lower on individualism ($-57.6$), masculinity ($-87.7$) and indulgence ($-49.6$), each in 6 of 6 pairs, with similar values for American personas. The age gap in individualism was as large as the national gap ($-73.6$ in English, $-54.1$ in Arabic), and the age gap in masculinity exceeded any national difference.

\subsection{Confidence}

\begin{figure*}[t]
\centering
\includegraphics[width=\textwidth]{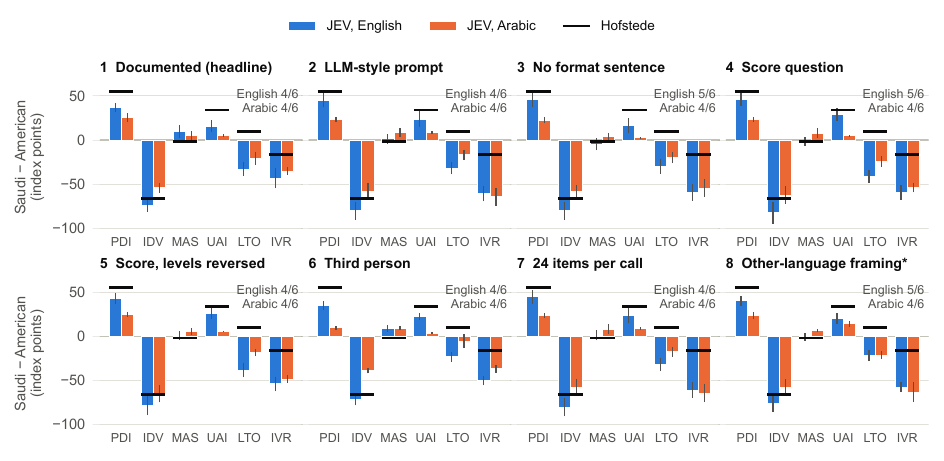}
\caption{Rank test by set-up. Bars: Saudi minus American persona on each dimension (means over the 12 matched profiles with 95\% CIs); black lines: the Hofstede difference. A bar on the other side of zero from its line marks a dimension on which JEV orders the two countries differently from Hofstede's scores. Top right: dimensions in Hofstede's order, of six. PDI: power distance, IDV: individualism, MAS: masculinity, UAI: uncertainty avoidance, LTO: long-term orientation, IVR: indulgence. *English and Arabic refer to the items; the framing is in the other language.}
\label{fig:rank}
\end{figure*}

In the documented set-up, JEV was less confident in Arabic, by 0.090 on average. The gap appeared for all 25 personas (Wilcoxon $p=6\times10^{-8}$) and 20 of 24 items ($p=1.1\times10^{-4}$), and its two-way cluster bootstrap interval, which resamples personas and items, ran from 0.048 to 0.132. Confidence for Saudi personas was lower by 0.024, a smaller and less consistent gap: it appeared for 18 of 24 matched profiles ($p=0.008$) but only 16 of 24 items ($p=0.11$), and its two-way interval ($-0.022$ to 0.067) included zero. Across set-ups, mean confidence was lower in Arabic in seven and lower for Saudi personas in all eight, but the two-way intervals excluded zero only in the documented and both Score set-ups for Arabic, and only in the two Score set-ups for Saudi personas (Appendix~\ref{app:setups}). In the language cross, Arabic items under English framing received slightly higher confidence than English items under Arabic framing: confidence tended to follow the language of the framing, whereas the national difference was associated mainly with the language of the items. Confidence was lowest for the agree--disagree items on managers (0.35) and on rules (0.38) and highest for how often one feels nervous or tense (0.75).

\subsection{Robustness to Request Design}
The set-up changed the size of the effects but not the direction of the national difference (Figure~\ref{fig:rank}; Appendix~\ref{app:setups}). The English share ranged from 87.1\% to 106.8\% and the Arabic share from 42.1\% (third person) to 72.2\%; the discrepancy ranged from 57.7\% to 73.5\% in English and from 59.9\% to 74.4\% in Arabic. In every set-up the Arabic share was smaller than the English share, whereas the Arabic discrepancy was larger in five set-ups and smaller in three. In every set-up the English difference pointed in the reference direction (cosine 0.82--0.85), female Saudi personas were more individualist and less masculine than male ones, and elder Saudi personas scored higher on power distance and lower on individualism than young ones. Sending the 24 items in one request left the results practically unchanged (English 103.8\% against 103.5\%, Arabic 69.9\% against 70.2\%).

\subsection{Rank Agreement}
In the rank test of \citet{masoud2025cultural}, JEV ordered Saudi Arabia and the United States as Hofstede's scores do on four of the six dimensions in both languages: power distance, individualism, uncertainty avoidance and indulgence, but not masculinity or long-term orientation (mean $\tau=0.33$; 33\% of dimensions mis-ranked). The profile of \citet{almutairi2021reclaiming} gives the same result, as its differences have the same signs. Across set-ups, four or five dimensions were in order in English and four in Arabic (Figure~\ref{fig:rank}; Table~\ref{tab:rank}). Long-term orientation was out of order in every set-up and language, and masculinity, on which Hofstede's scores differ by 2 points, in all eight set-ups in Arabic and in five in English. No matched profile had all six dimensions in order in any set-up. Without a persona, the language alone did not produce the national order: the Arabic profile differed from the English one in the reference direction on one dimension in the documented set-up (individualism; $\tau=-0.67$) and on three or four in the others.

\section{Discussion}

\subsection{Cultural Patterns without Text Generation}
JEV shows patterns that have been reported for generative models: a default nearer to American than to Saudi personas, partial movement towards a named country, a weaker national difference in a language other than English, and stereotyped answers for demographic groups \citep{durmus2023towards,tao2024cultural,alkhamissi2024investigating,wang2024not}. In JEV these patterns arise without decoding, sampling or refusals. This suggests that they reside in what the model has learned about groups, languages and survey items rather than in the generation of text, and that replacing generation with calibrated decisions does not by itself remove them.

\subsection{Direction Is Not Accuracy}
The share and the discrepancy answer different questions. In English, JEV moved most of the way along the reference direction, but its national difference also pointed elsewhere, most clearly in long-term orientation and in an exaggerated indulgence gap, so that it departed from the reference by about 60\% of the reference length; a model with no national difference at all would depart by 100\%. In Arabic the share fell but the discrepancy did not, because the deviations shrank with the difference. A stand-in for respondents needs both, so evaluations should report how far the whole profile of differences departs from a benchmark, not only how far it moves towards it. Because the benchmark itself comes from unmatched samples and periods, neither measure says how well JEV matches Saudi or American people today.

\subsection{Language: Mainly the Items}
In this experiment, the language cross associated the Arabic effect mainly with the item wording ($-29.5$ points) rather than with the language of the persona description ($-3.9$, an interval that does not exclude a small effect). The largest item-level losses were in agree--disagree statements about organisational hierarchy and rules, which suggests that the Arabic formulations of these statements carry less of the national contrast than their English counterparts. For multilingual evaluation this means that the translation of the items, not only the language of the prompt, is part of what is measured. With a single translation, however, we cannot tell whether the effect reflects Arabic as such or the choices of this translation. Nor did the language stand in for the culture: without a persona, Arabic items did not move JEV towards the Saudi order, a caution for comparisons that let the prompt language represent a country \citep{masoud2025cultural}.

\subsection{Personas as Population Proxies}
Age shifted JEV's profiles about as much as nationality, gender shifted them in stereotyped directions, and each added attribute diluted the national difference (in English from 96\% for country-only personas to 85\% with both gender and age, in Arabic from 81\% to 57\%). Population experiments that condition JEV on several demographic attributes \citep{li2026kite} should therefore expect the national signal to weaken as attributes are added, and age to carry stereotyped contrasts, such as a masculinity gap between elder and young personas larger than any national difference. The larger gender contrast for Saudi than for American personas is a further stereotype that simulated Saudi samples would inherit.

\subsection{Confidence as a Signal}
In the documented set-up, JEV was less confident in Arabic, and its confidence tended to follow the language of the framing rather than of the items; lower confidence for Saudi personas was small and held consistently only in the Score set-ups. Lower confidence in Arabic could help to flag answers that rest on less evidence. It does not show that the model is calibrated to the diversity of real respondents: a value question has no correct option, and confidence measures the model's certainty, not the agreement among people. Testing whether JEV's probabilities match the answer distributions of Saudi respondents, for example in the World Values Survey or the Arab Barometer, is the natural next step.

\section{Conclusion}

We audited the cultural values of a decision-only model with the VSM 2013, matched Saudi and American personas, two languages and eight request designs. JEV's answers are highly repeatable, nearer to its American than to its Saudi personas when no persona is given, and responsive to nationality. In English they cover most of the reference Saudi--US difference along its direction, and in Arabic less, mainly in association with the item wording rather than the language of the persona; in both languages, the full profile of differences departs from the reference by about 60\% of its size. Age and gender shift the profiles in stereotyped ways, and in the documented design confidence is lower in Arabic. The direction of the national difference holds across request designs. A model that never writes text is thus not free of the cultural patterns known from generative models, and using it as a stand-in for respondents calls for the same scrutiny.

\section*{Limitations}

We studied one version of one model, whose training is not public. Our Arabic results rest on a single official translation, so item-language effects cannot be separated from its wording. The Hofstede scores come from unmatched samples, periods and survey versions, partly from extensions or estimates \citep{hofstede2013vsm}; they are a benchmark for the direction and size of a two-country difference, not ground truth or a picture of opinion within countries, and the two-country rank test treats near-ties like large differences. Our intervals cover only 12 constructed persona profiles and 24 items, not real populations, other persona wordings or translations, or error in the reference scores. The analysis was exploratory and not pre-registered, and other request formulations may give other effect sizes.

\section*{Ethical Considerations}

The study involves no human participants. Asking a model to answer as members of national, gender and age groups elicits stereotyped portrayals; we report them to document the risk, not to endorse them, and we caution against using such outputs as substitutes for the views of the groups concerned. The prompts, all responses and the analysis code will be released to allow verification.

\bibliography{references}

\appendix
\section{VSM 2013 Scoring and Reference Values}\label{app:vsm}

The six indices are the following sums, each plus a constant $C$ chosen by the researcher \citep{hofstede2013vsm}:
{\small
\begin{align*}
\text{PDI} &= 35(m_{7}-m_{2}) + 25(m_{20}-m_{23}),\\
\text{IDV} &= 35(m_{4}-m_{1}) + 35(m_{9}-m_{6}),\\
\text{MAS} &= 35(m_{5}-m_{3}) + 35(m_{8}-m_{10}),\\
\text{UAI} &= 40(m_{18}-m_{15}) + 25(m_{21}-m_{24}),\\
\text{LTO} &= 40(m_{13}-m_{14}) + 25(m_{19}-m_{22}),\\
\text{IVR} &= 35(m_{12}-m_{11}) + 40(m_{17}-m_{16}),
\end{align*}}
where $m_k$ is the mean answer to item $k$. Every measure in this paper is a difference between two profiles or a ratio of such differences, so the constants $C$ cancel. Table~\ref{tab:reference} lists the reference scores.

\begin{table}[h]
\centering\small\setlength{\tabcolsep}{4pt}
\begin{tabular}{@{}lccccc@{}}
\toprule
 & \multicolumn{2}{c}{Hofstede} & Alm. & \multicolumn{2}{c}{Saudi $-$ US} \\
\cmidrule(lr){2-3}\cmidrule(lr){5-6}
Dimension & SA & US & SA & Hof. & Alm. \\
\midrule
Power distance & 95 & 40 & 72 & $+$55 & $+$32 \\
Individualism & 25 & 91 & 48 & $-$66 & $-$43 \\
Masculinity & 60 & 62 & 43 & $-$2 & $-$19 \\
Uncertainty avoidance & 80 & 46 & 64 & $+$34 & $+$18 \\
Long-term orientation & 36 & 26 & 27 & $+$10 & $+$1 \\
Indulgence & 52 & 68 & 14 & $-$16 & $-$54 \\
\bottomrule
\end{tabular}
\caption{Reference scores: Hofstede's profiles for Saudi Arabia (SA; \citealp{hofstedeinsights2022}) and the United States (US; \citealp{hofstede2010cultures}), the Saudi profile of \citet{almutairi2021reclaiming} (Alm.), and the reference Saudi-minus-US differences.}
\label{tab:reference}
\end{table}

\section{Prompts and Requests}\label{app:prompts}

The English texts are given in Section~\ref{sec:setups} and Table~\ref{tab:setups}. Table~\ref{tab:arabic-framing} gives the Arabic wording of the persona prompt and of the texts written for set-ups 1 and 6, and Figure~\ref{fig:request} shows a complete request in the documented set-up.

\begin{table*}[h]
\centering
\includegraphics[width=\textwidth]{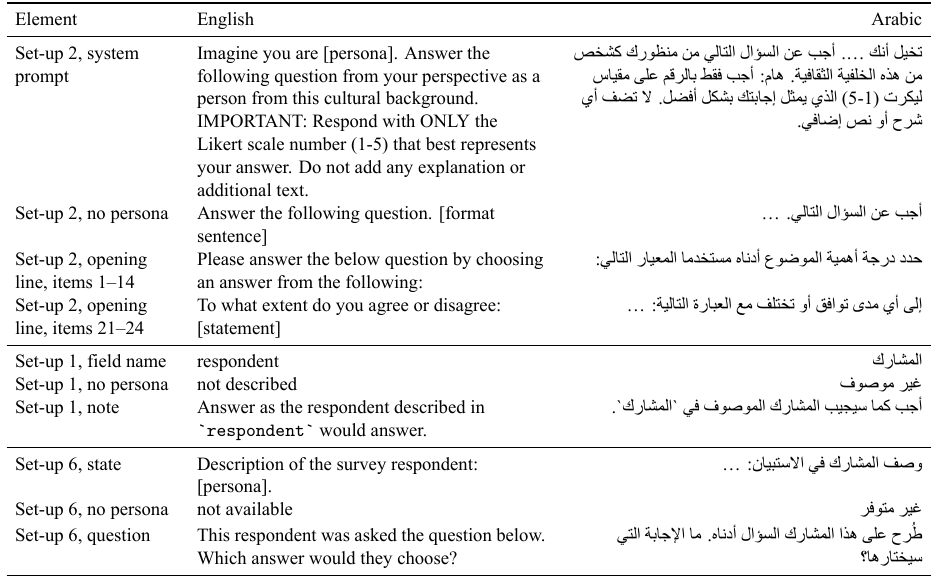}
\caption{English and Arabic wording of the persona prompt (set-up 2) and of the texts written for set-ups 1 and 6. Brackets and ellipses mark where the persona description, the format sentence or the statement is inserted; the Arabic prompts use Arabic persona descriptions.}
\label{tab:arabic-framing}
\end{table*}

\begin{figure*}[h]
\centering
\begin{minipage}{0.86\textwidth}
\small
\begin{verbatim}
POST https://api.typesafe.ai/v1/systemone
{
  "model": "jev-1.13.0",
  "state": {"respondent": "a young female from Saudi Arabia"},
  "questions": {"answer": {
    "type": "choice",
    "instructions": {
      "question": "In choosing an ideal job, how important would it be to you
                   to have sufficient time for your personal or home life?",
      "note": "Answer as the respondent described in `respondent` would answer."},
    "criteria": {"1": "of utmost importance", "2": "very important",
                 "3": "of moderate importance", "4": "of little importance",
                 "5": "of very little or no importance"}}}
}
\end{verbatim}
\end{minipage}
\caption{A request in the documented set-up (item 1, English). The response gives the most probable option, a probability for each option and a confidence value.}
\label{fig:request}
\end{figure*}

\section{Reliability by Set-up}\label{app:reliability}

Table~\ref{tab:reliability} reports test--retest reliability and the shape of the answer distributions for each set-up.

\begin{table*}[h]
\centering\small
\begin{tabular}{@{}lccccc@{}}
\toprule
Set-up & ICC & Same top option (30 calls) & SD across calls & Mean confidence & Normalised entropy \\
\midrule
Documented (headline) & 0.997 & 85.0\% & 0.031 & 0.49 & 0.61 \\
LLM-style prompt & 0.997 & 88.7\% & 0.030 & 0.54 & 0.55 \\
No format sentence & 0.998 & 89.3\% & 0.026 & 0.58 & 0.49 \\
Score question & 0.996 & 85.4\% & 0.032 & 0.55 & 0.66 \\
Score, levels reversed & 0.995 & 83.3\% & 0.032 & 0.52 & 0.69 \\
Third person & 0.997 & 87.4\% & 0.030 & 0.58 & 0.49 \\
24 items per call & 0.997 & 88.4\% & 0.030 & 0.54 & 0.55 \\
Other-language framing* & 0.997 & 85.7\% & 0.029 & 0.54 & 0.56 \\
\bottomrule
\end{tabular}
\caption{Test--retest reliability over the 30 calls of each prompt and the shape of JEV's answer distributions, by set-up (1,200 prompts each). *Framing and items in different languages.}
\label{tab:reliability}
\end{table*}

\section{Findings by Set-up}\label{app:setups}

Table~\ref{tab:setup-findings} reports the share of the reference difference, the discrepancy, the default position and the confidence gaps for each set-up, Table~\ref{tab:default} the distances of the no-persona profile from the persona centroids, and Table~\ref{tab:rank} the rank test.

\begin{table*}[h]
\centering\small\setlength{\tabcolsep}{4pt}
\begin{tabular}{@{}lcccccccc@{}}
\toprule
 & \multicolumn{2}{c}{Share (\%)} & \multicolumn{2}{c}{Discrepancy (\%)} & \multicolumn{2}{c}{Default position} & \multicolumn{2}{c}{Confidence gap} \\
\cmidrule(lr){2-3}\cmidrule(lr){4-5}\cmidrule(lr){6-7}\cmidrule(lr){8-9}
Set-up & English & Arabic & English & Arabic & English & Arabic & EN $-$ AR & Am.\ $-$ Saudi \\
\midrule
Documented (headline) & 87.1 & 62.0 & 62.3 & 59.9 & 0.08 & 0.06 & $+$0.090$^{\dagger}$ & $+$0.024 \\
LLM-style prompt & 103.5 & 70.2 & 68.2 & 74.1 & $-$0.25 & 0.02 & $+$0.023 & $+$0.020 \\
No format sentence & 101.3 & 65.2 & 68.0 & 71.3 & $-$0.17 & $-$0.08 & $+$0.008 & $+$0.018 \\
Score question & 106.8 & 69.4 & 73.5 & 71.8 & $-$0.04 & $-$0.00 & $+$0.063$^{\dagger}$ & $+$0.035$^{\dagger}$ \\
Score, levels reversed & 99.8 & 72.2 & 68.1 & 63.9 & $-$0.05 & $-$0.01 & $+$0.090$^{\dagger}$ & $+$0.040$^{\dagger}$ \\
Third person & 89.1 & 42.1 & 57.7 & 71.2 & 0.13 & 0.53 & $+$0.028 & $+$0.028 \\
24 items per call & 103.8 & 69.9 & 68.4 & 74.4 & $-$0.26 & 0.03 & $+$0.023 & $+$0.020 \\
Other-language framing* & 97.2 & 71.6 & 60.2 & 72.6 & $-$0.12 & $-$0.03 & $-$0.028 & $+$0.008 \\
\bottomrule
\end{tabular}
\caption{Main findings by set-up. Share: of the reference (Hofstede) Saudi--US difference, along its direction. Discrepancy: length of JEV's difference minus the reference difference, relative to the reference length. Default position: 0 = JEV's American personas, 1 = its Saudi personas. Confidence gaps: mean differences over matched prompts; $^{\dagger}$ the two-way cluster bootstrap interval (personas and items) excludes zero. *English and Arabic refer to the items; the framing is in the other language.}
\label{tab:setup-findings}
\end{table*}

\begin{table*}[h]
\centering\small\setlength{\tabcolsep}{4pt}
\begin{tabular}{@{}lcccccccc@{}}
\toprule
 & \multicolumn{4}{c}{English} & \multicolumn{4}{c}{Arabic} \\
\cmidrule(lr){2-5}\cmidrule(lr){6-9}
Set-up & to Am. & to Saudi & off axis & axis & to Am. & to Saudi & off axis & axis \\
\midrule
Documented (headline) & 61.5 & 111.0 & 61.0 & 100.3 & 54.5 & 87.5 & 54.3 & 73.0 \\
LLM-style prompt & 46.5 & 150.4 & 36.3 & 116.9 & 28.1 & 94.3 & 28.0 & 92.0 \\
No format sentence & 65.3 & 148.6 & 62.2 & 115.1 & 34.2 & 98.1 & 33.5 & 85.0 \\
Score question & 34.8 & 132.1 & 34.4 & 122.1 & 33.4 & 95.9 & 33.4 & 89.7 \\
Score, levels reversed & 38.7 & 125.8 & 38.2 & 113.9 & 25.9 & 91.6 & 25.9 & 87.1 \\
Third person & 49.1 & 99.2 & 47.5 & 99.6 & 50.1 & 48.3 & 40.6 & 55.7 \\
24 items per call & 47.5 & 151.9 & 36.6 & 117.2 & 26.7 & 93.4 & 26.6 & 92.0 \\
Other-language framing* & 49.8 & 129.6 & 48.2 & 107.8 & 35.3 & 101.7 & 35.1 & 92.4 \\
\bottomrule
\end{tabular}
\caption{The no-persona profile by set-up: its distances, in index points, to the centroids of JEV's American (Am.) and Saudi personas, its distance from the axis through the two centroids, and the length of that axis. *English and Arabic refer to the items; the framing is in the other language.}
\label{tab:default}
\end{table*}

\begin{table*}[h]
\centering\small\setlength{\tabcolsep}{4pt}
\begin{tabular}{@{}lcccccc@{}}
\toprule
 & \multicolumn{2}{c}{Personas, English} & \multicolumn{2}{c}{Personas, Arabic} & \multicolumn{2}{c}{No persona, Arabic $-$ English} \\
\cmidrule(lr){2-3}\cmidrule(lr){4-5}\cmidrule(lr){6-7}
Set-up & In order & Out of order & In order & Out of order & In order & Out of order \\
\midrule
Documented (headline) & 4 (0.33) & MAS, LTO & 4 (0.33) & MAS, LTO & 1 ($-$0.67) & PDI, MAS, UAI, LTO, IVR \\
LLM-style prompt & 4 (0.33) & MAS, LTO & 4 (0.33) & MAS, LTO & 3 (0.00) & MAS, LTO, IVR \\
No format sentence & 5 (0.67) & LTO & 4 (0.33) & MAS, LTO & 3 (0.00) & MAS, LTO, IVR \\
Score question & 5 (0.67) & LTO & 4 (0.33) & MAS, LTO & 4 (0.33) & LTO, IVR \\
Score, levels reversed & 4 (0.33) & MAS, LTO & 4 (0.33) & MAS, LTO & 4 (0.33) & LTO, IVR \\
Third person & 4 (0.33) & MAS, LTO & 4 (0.33) & MAS, LTO & 4 (0.33) & LTO, IVR \\
24 items per call & 4 (0.33) & MAS, LTO & 4 (0.33) & MAS, LTO & 3 (0.00) & MAS, LTO, IVR \\
Other-language framing* & 5 (0.67) & LTO & 4 (0.33) & MAS, LTO & 4 (0.33) & MAS, IVR \\
\bottomrule
\end{tabular}
\caption{Rank test of \citet{masoud2025cultural} by set-up: number of dimensions (of six) on which JEV orders Saudi Arabia and the United States as Hofstede's scores do, with the mean Kendall's $\tau$ in parentheses, and the dimensions out of order. Personas: Saudi minus American personas (means over the 12 matched profiles). No persona: the Arabic minus the English no-persona profile, with the language standing for the country. The Saudi profile of \citet{almutairi2021reclaiming} gives the same orders. *English and Arabic refer to the items; the framing is in the other language.}
\label{tab:rank}
\end{table*}

\section{Additional Results}\label{app:items}

Table~\ref{tab:dims-main} gives the Saudi--American differences by dimension for the LLM-style prompt, and Table~\ref{tab:items} the item-level differences and confidence in the documented set-up.

\begin{table*}[h]
\centering\small
\begin{tabular}{@{}lccc@{}}
\toprule
Dimension & Hofstede & English & Arabic \\
\midrule
Power distance & $+$55 & $+$44.9 [36.9, 53.0] & $+$23.4 [20.4, 26.4] \\
Individualism & $-$66 & $-$80.0 [$-$90.3, $-$69.7] & $-$57.9 [$-$66.9, $-$48.9] \\
Masculinity & $-$2 & $+$1.4 [$-$4.3, 7.2] & $+$8.4 [3.1, 13.7] \\
Uncertainty avoidance & $+$34 & $+$23.7 [14.7, 32.7] & $+$8.4 [6.9, 10.0] \\
Long-term orientation & $+$10 & $-$31.7 [$-$38.8, $-$24.6] & $-$16.7 [$-$22.6, $-$10.9] \\
Indulgence & $-$16 & $-$60.6 [$-$69.5, $-$51.7] & $-$64.4 [$-$74.2, $-$54.5] \\
\bottomrule
\end{tabular}
\caption{Saudi minus American persona by dimension with the LLM-style prompt (means over 12 matched profiles, 95\% CIs), with the Hofstede difference.}
\label{tab:dims-main}
\end{table*}

\begin{table*}[h]
\centering\small
\begin{tabular}{@{}rllcccc@{}}
\toprule
 & & & \multicolumn{2}{c}{Saudi $-$ American answer} & \multicolumn{2}{c}{Confidence} \\
\cmidrule(lr){4-5}\cmidrule(lr){6-7}
Item & Dim. & Content & English & Arabic & English & Arabic \\
\midrule
1 & IDV & time for personal or home life & $+$0.01 & $-$0.06 & 0.59 & 0.42 \\
2 & PDI & a boss you can respect & $-$0.22 & $-$0.51 & 0.59 & 0.35 \\
3 & MAS & recognition for good performance & $-$0.09 & $-$0.11 & 0.43 & 0.41 \\
4 & IDV & security of employment & $-$0.44 & $-$0.34 & 0.55 & 0.44 \\
5 & MAS & pleasant people to work with & $-$0.07 & $-$0.12 & 0.68 & 0.41 \\
6 & IDV & interesting work & $+$0.53 & $+$0.23 & 0.49 & 0.30 \\
7 & PDI & being consulted by your boss & $+$0.20 & $-$0.34 & 0.47 & 0.32 \\
8 & MAS & living in a desirable area & $+$0.09 & $-$0.00 & 0.41 & 0.40 \\
9 & IDV & a job respected by family and friends & $-$1.13 & $-$1.03 & 0.54 & 0.42 \\
10 & MAS & chances for promotion & $-$0.17 & $-$0.17 & 0.54 & 0.44 \\
11 & IVR & keeping time free for fun & $+$0.39 & $+$0.14 & 0.53 & 0.44 \\
12 & IVR & moderation: having few desires & $-$0.69 & $-$0.69 & 0.57 & 0.49 \\
13 & LTO & doing a service to a friend & $-$0.48 & $-$0.48 & 0.55 & 0.41 \\
14 & LTO & thrift & $-$0.07 & $-$0.40 & 0.57 & 0.49 \\
15 & UAI & how often nervous or tense & $+$0.13 & $+$0.06 & 0.79 & 0.70 \\
16 & IVR & are you a happy person & $+$0.06 & $+$0.04 & 0.51 & 0.59 \\
17 & IVR & others prevent you from doing what you want & $-$0.08 & $-$0.10 & 0.57 & 0.64 \\
18 & UAI & state of health & $-$0.01 & $+$0.00 & 0.57 & 0.61 \\
19 & LTO & national pride & $-$0.77 & $-$0.84 & 0.58 & 0.50 \\
20 & PDI & subordinates afraid to contradict the boss & $+$0.57 & $+$0.81 & 0.62 & 0.38 \\
21 & UAI & good manager without precise answers & $+$0.29 & $+$0.24 & 0.41 & 0.29 \\
22 & LTO & persistent efforts give results & $-$0.10 & $-$0.14 & 0.45 & 0.36 \\
23 & PDI & avoid structures with two bosses & $-$0.30 & $+$0.00 & 0.38 & 0.44 \\
24 & UAI & rules should not be broken & $-$0.54 & $-$0.06 & 0.39 & 0.37 \\
\bottomrule
\end{tabular}
\caption{Item-level results in the documented set-up: mean difference between Saudi and American personas in the expected answer (1--5 scale; negative = Saudi personas chose options nearer the first, e.g.\ more important or prouder), and JEV's mean confidence, by language.}
\label{tab:items}
\end{table*}

\section{Cost, Software and Reproducibility}\label{app:repro}

The 288,000 answers took 253,500 requests (1,500 in the batched set-up) and 131 million input tokens, at a cost of about US\$5.50 (US\$0.042 per million input tokens; output tokens are free). Requests were sent with up to 16 in parallel per set-up, and server errors were retried. The analysis used Python 3.12 with NumPy 1.26, pandas 2.2, SciPy 1.16 and matplotlib 3.10, with a fixed seed for bootstrap resampling. The prompts, all responses and the analysis code will be released.

\end{document}